\documentclass[preprint]{elsarticle}

\usepackage{wrapfig}

\usepackage{mathtools}
\usepackage{graphicx}   
\usepackage{color}
\usepackage{ulem}
\usepackage{enumitem}
\setlist{leftmargin=4mm}
\usepackage{setspace}
\usepackage[margin=1.in]{geometry}

\usepackage{caption} 
\usepackage{subcaption}  
\usepackage{amsfonts}

\usepackage{algorithm}
\usepackage{algpseudocode}
\usepackage{algorithmicx}

\usepackage{etoolbox} 
\makeatletter
\patchcmd{\subsubsection}{\itshape}{\bfseries}{}{}
\makeatother

\usepackage{etoolbox}
\patchcmd{\section}
  {\centering}
  {\raggedright}
  {}
  {}
\patchcmd{\section}
  {\centering}
  {\raggedright}
  {}
  {}
\patchcmd{\subsubsection}
  {\centering}
  {\raggedright}
  {}
  {}
\usepackage{float}

\makeatletter

\renewcommand*{\p@subsection}{}

\renewcommand*{\p@subsubsection}{}
\makeatother

\usepackage{amsmath}

\usepackage{mathtools}
\usepackage{bm}

\usepackage{soul}

\usepackage{etoolbox}


\usepackage{array}
\usepackage{booktabs}
\usepackage{multirow}

\usepackage{booktabs}

\begin{document}

\singlespacing

\begin{frontmatter}
\title{Mitigating mode collapse in normalizing flows by annealing with an adaptive schedule:  Application to parameter estimation}

\author[1,2]{Yihang Wang}
\author[1,2]{Chris Chi}
\author[1,2]{Aaron R. Dinner}
\ead{dinner@uchicago.edu}

\affiliation[1]{
organization={Department of Chemistry, University of Chicago}, 
city={Chicago}, 
state={Illinois}, 
postcode={60637}, 
country={United States}}

\affiliation[2]{organization={James Franck Institute, University of Chicago}, city={Chicago}, 
state={Illinois}, 
postcode={60637}, 
country={United States}}

\begin{abstract} 
Normalizing flows (NFs) provide uncorrelated samples from complex distributions, making them an appealing tool for parameter estimation. However, the practical utility of NFs remains limited by their tendency to collapse to a single mode of a multimodal distribution. In this study, we show that  annealing with an adaptive schedule based on the effective sample size (ESS) can mitigate mode collapse. We demonstrate that our approach can converge the marginal likelihood for a biochemical oscillator model fit to time-series data in ten-fold less computation time than a widely used ensemble Markov chain Monte Carlo (MCMC) method. We show that the ESS can also be used to reduce variance by pruning the samples.  We expect these developments to be of general use for sampling with NFs and discuss potential opportunities for further improvements.
\end{abstract}

\begin{keyword}
Bayesian inference, normalizing flows, adaptive annealing, mode collapse
\end{keyword}

\end{frontmatter}

\section{\large{Introduction}}

Fitting models to data enables the objective evaluation of the models for interpretation of the data and estimation of parameters; in turn, models can be used for both interpolation of the data and extrapolation for prediction. Because few models are analytically tractable, fitting typically relies on sampling parameters through Markov chain Monte Carlo (MCMC) simulations, in which new values for parameters are proposed and accepted or rejected so as to sample a desired distribution in the limit of many proposals. Unfortunately, MCMC simulations often converge slowly.  Proposals are generally incremental changes to the parameter values, and correlations between parameters cause distributions to be strongly anisotropic \cite{universally_sloppy, James_Sloppiness}---limiting the sizes of the changes---and multimodal---requiring traversal of low probability regions.

In the case that we consider here---fitting time series data with systems of ordinary differential equations (ODEs)---these issues are exacerbated by separations of time scales between the time steps for numerical integration, which are set by the fastest processes described, and the total times of integration, which are set by the slowest processes of interest.  Such separations of time scales make evaluating proposals and, when gradients are needed, making proposals computationally costly.  This limits the number of proposals that are computationally feasible and, in turn, the use of many methods developed to address the anisotropy and multimodality \cite{richardson1997bayesian,goodman2010ensemble,chopin2012free,foreman2013emcee,matthews2018umbrella,dinner2020stratification,Grant_NF_MCMC,chi2024sampling}.

Given recent rapid advances in machine learning, researchers have sought to use it to accelerate sampling \cite{ML_enhanced_MD, mehdi_review}. The main idea of the methods that have been proposed is to learn the structure of the target distribution and use this information to guide sampling.  One method that is suitable for this purpose is to learn a normalizing flow (NF), which is an invertible map from a tractable (base) distribution to the target distribution \cite{NF, NF2020review}. If such a map can be learned, uncorrelated samples with high likelihoods of acceptance can be generated rapidly by drawing independently from the base distribution and applying the map. 
Because NFs offer an exact calculation of the probability density of samples, they can be combined with different statistical estimators to provide unbiased estimates of expectation values, even when they do not match target distributions exactly \cite{nicoli2020asymptotically}. 

For example, No\'{e} {\itshape et al.} showed that an NF can be used to generate low-energy conformations of condensed phase systems and sample metastable states to estimate their relative free energies \cite{Boltzmann_generators} (see \cite{coretti2024boltzmann} and references therein for commentary and related work). Of special interest for our study, Gabri\'{e} {\itshape et al.} investigated the use of NFs to accelerate sampling a Bayesian posterior distribution for parameters of a model similar to one for a star-exoplanet system \cite{gabrie2021efficient}; in that study and a follow-on one \cite{Grant_NF_MCMC}, they showed that combining conventional Metropolis-Hastings MCMC sampling with proposals from NFs enhanced sampling efficiency. Grumitt {\itshape et al.} introduced a deterministic Langevin Monte Carlo algorithm that replaces the stochastic term in the Langevin equation with a deterministic density-gradient term evaluated using NFs and demonstrated that it improved sampling distributions representative of ones encountered in Bayesian parameter estimation, particularly for cases where direct calculation of the likelihood gradient is computationally expensive \cite{LangevinMC_NF}. Souveton {\it et al.} used flows based on Hamiltonian dynamics to sample the posterior distribution
of parameters of a cosmological model \cite{souveton2024fixed}.

Despite these successes, a significant challenge in using NFs is their potential for mode collapse, in which the model tends to focus on a single mode of a multimodal target distribution \cite{mode-collapse_NF,Cranmer2021flow_lattice_filed}.  Gabri\'{e} {\itshape et al.} assume that the modes are known so that samples can be initialized in each, and in some cases the modes can be anticipated from symmetry \cite{Cranmer2021flow_lattice_filed}. However,  often such information is not available a priori. Various approaches to mitigate mode collapse have been investigated. As we discuss further below, an NF is trained by minimizing a divergence (typically, the Kullback-Leibler divergence) between the distribution produced by the model and a target distribution, and the choice of divergence can promote or suppress mode collapse \cite{Cranmer2021flow_lattice_filed,mode-collapse_NF,mate2023learning}. 
Researchers have also investigated tuning the transport from the base to the target distributions \cite{mate2023learning}, alternative gradient estimators \cite{vaitl2022gradients}, and annealed importance sampling between the model and target distributions \cite{midgley2023flow}.  Ultimately, mitigating mode collapse is likely to require a combination of approaches, and it is important to test the effectiveness of approaches on different classes of problems.

In this study, we explore mitigating mode collapse in normalizing flows for parameter estimation in a Bayesian framework by annealing from the prior distribution to the posterior distribution.  Specifically, we show how the effective sample size can be used (1) to determine an adaptive annealing schedule to sample a multimodal parameter distribution for a model of a biochemical oscillator robustly without prior knowledge of the modes and (2) to prune the results to reduce the variance.  For hyperparameter values tested, we are able to achieve a ten-fold speedup relative to a widely used Markov chain Monte Carlo (MCMC) ensemble sampler.  Potential directions for future research are discussed.\\

\section{\large{Normalizing flows}}

Using NFs to generate MC moves presents a fundamental dilemma: the NF must learn the structure of the target distribution from the data, yet the data are only obtained once the parameter space is explored. Below, we describe a training scheme that allows the NF to explore the target space autonomously, followed by a discussion of strategies to enhance the robustness and reliability of this scheme.

\subsection{Architecture}

An NF transforms a sample $\mathbf{z}$ from the base distribution, $p_z(\mathbf{z})$ into a sample  $\mathbf{x}$ from an approximation $q_\phi(\mathbf{x})$ to the target distribution $p(\mathbf{x})$ through a sequence of invertible, differentiable functions $\{f_i\}_{i=1}^N$ with learnable parameters $\phi$:
\begin{equation}
    \mathbf{x} =f_\phi ( \mathbf{z})= f_N \circ f_{N-1} \circ \ldots \circ f_1(\mathbf{z}).
\end{equation}
By the change of variables theorem, 
\begin{equation}
    q_\phi(\mathbf{x})  = p_z(f_\phi^{-1}(\mathbf{x})) \left| \det \left( \frac{d f_\phi^{-1}}{d \mathbf{x}} \right) \right|,
\end{equation}
where $f_\phi^{-1}$ is the inverse of the composite function and $d f_\phi^{-1}/d \mathbf{x}$ is its Jacobian. 
In practice, it is important to choose a form for $f_i$ that facilitates computing the determinant of the Jacobian; we use RealNVP (Non-Volume Preserving) \cite{RealNVP} in our numerical example. 

The RealNVP architecture consists of $L$ layers. In each layer $\ell$, the input with dimension $V$ is split into two parts, each with dimension $v= V/2$. The first half (denoted by $x^\ell_{1: v}$) remains unchanged, and the second half (denoted by $x^\ell_{v+1: V}$) is subject to an affine transformation based on the first part:
\begin{equation}
\begin{aligned}
& \mathbf{x}^{\ell+1}_{1: v} =\mathbf{x}^\ell_{1: v} \\
& \mathbf{x}^{\ell+1}_{v+1: V} =\mathbf{x}^\ell_{v+1: V} \odot \exp \left(a^\ell_\phi\left(\mathbf{x}^\ell_{1: v}\right)\right)+b^\ell_\phi\left(\mathbf{x}^\ell_{1: v}\right),
\end{aligned}
\label{eq.realNVP}
\end{equation}
where $\odot$ represents element-wise multiplication, and $a^\ell_\phi$ and $b^\ell_\phi$ are learnable scaling and translation parameters that are represented by neural networks. Owing to this structure, the Jacobian matrix is triangular, and the determinant is a product of diagonal elements:
\begin{equation}
    \left| \det \left( \frac{d \mathbf{x}^{\ell+1}}{d \mathbf{x}^\ell} \right) \right| = \prod_{i=v+1}^{V} \exp\left(a^\ell_\phi(\mathbf{x}_{1:v}^\ell)_i\right).
\end{equation}

\subsection{Training}

To train the NF, we minimize the Kullback-Leibler (KL) divergence between the approximation $q_{\phi}(\mathbf{x})$ and the target distribution $p(\mathbf{x})$. Because the KL divergence is nonsymmetric with respect to its arguments, there are two possible loss functions \cite{NF}.  The ``reverse'' loss is
\begin{equation} \label{eqn:reverse}
{\cal L} = \mathrm{KL}\left(q_\phi \| p\right)=\int q_\phi(\mathbf{x}) \ln \frac{q_\phi(\mathbf{x})}{p(\mathbf{x})}d\mathbf{x}=\left\langle \ln \frac{q_\phi(\mathbf{x})}{p(\mathbf{x})}\right\rangle_{q_\phi},
\end{equation}
and the ``forward'' loss is
\begin{equation}
{\cal L}  = \mathrm{KL}\left(p \| q_\phi\right)=\int  p(\mathbf{x}) \ln \frac{p(\mathbf{x})}{q_\phi(\mathbf{x})}d\mathbf{x}=\left\langle \ln \frac{p(\mathbf{x})}{q_\phi(\mathbf{x})}\right\rangle_{p},
\end{equation}
where $\langle\ldots\rangle_p$ denotes an expectation over distribution $p$.  Despite their similarity, these two losses result in different performance. 

As the second equality in \eqref{eqn:reverse} indicates, the reverse loss can be viewed as an expectation over $q_\phi$, so that it can be evaluated by drawing samples from the model.  This feature is attractive because drawing samples from the model is computationally inexpensive compared with generating uncorrelated samples by MCMC for applications that we expect NFs to accelerate (in the case of parameter estimation, ones in which each evaluation of the likelihood is computationally expensive). However, consistent with previous observations \cite{mode-collapse_NF,Grant_NF_MCMC,mate2023learning}, we find that training with the reverse loss is prone to mode collapse (often termed ``mode-seeking'').  Not only does the reverse loss fail to penalize errors in regions of the space of interest where $q_\phi$ is small owing to the factor of $q_\phi$ in the integral, but $\ln q_\phi/p$ is large (i.e., unfavorable) where $q_\phi \gg p$, which penalizes  extension of the model tails beyond those of the target distribution.



The forward loss does not suffer from these issues, but, as written, it requires generating samples from the target distribution (e.g., by MCMC), which can be computationally costly. This issue can be overcome by importance sampling. That is, data can be drawn from an NF model $q_{\phi'}(\mathbf{x})$ and reweighted \cite{Cranmer2021flow_lattice_filed}:
\begin{equation}\label{eqn:forward_mode}
{\cal L}  = \mathrm{KL}\left(p \| q_\phi\right)=\int  q_{\phi'}(\mathbf{x}) \frac{ p(\mathbf{x})}{ q_{\phi'}(\mathbf{x})}\ln \frac{p(\mathbf{x})}{q_\phi(\mathbf{x})} d\mathbf{x} =\left\langle -\frac{ p(\mathbf{x})}{ q_{\phi'}(\mathbf{x})}  \ln {q_\phi(\mathbf{x})}\right\rangle_{q_{\phi'}}+C,
\end{equation}
where $C = \left\langle \ln p(\mathbf{x}) \right\rangle_{ p(\mathbf{x})}$ is a constant. In this case, the forward loss can again fail to penalize errors in regions of the space of interest where $q_\phi$ is small, but the ratio $p/q_\phi$ is large where $p \gg q_\phi$, which favors extension of the model tails beyond those of the target distribution (in this sense, training with the forward loss is ``mode-covering'').

The reverse and forward losses can be combined with each other as well as with MCMC \cite{mode-collapse_NF,Grant_NF_MCMC}. In our preliminary experiments, we explored many such possibilities, but we did not find for our application that combining approaches yielded better results than those obtained  training with only the forward loss in the importance sampling form in \eqref{eqn:forward_mode}.  We thus present results only for the latter approach.

For training, we require the gradient of ${\cal L}$ with respect to the parameters $\phi$.  In practice, it can be estimated as:
\begin{align}
\frac{\partial {\cal L}}{\partial \phi} 
&= \int  p(\mathbf{x}) \frac{\partial}{\partial \phi} \left[ \ln \frac{p(\mathbf{x})}{q_\phi(\mathbf{x})} \right] d\mathbf{x}  \nonumber \\
&= -\int  q_{\phi'}(\mathbf{x}) \frac{p(\mathbf{x})}{q_{\phi'}(\mathbf{x})} \frac{\partial}{\partial \phi} \left[ \ln q_\phi(\mathbf{x})  \right]d\mathbf{x} \nonumber \\
&\approx -\left. \sum_{\mathbf{x}\sim q_{\phi'}} w(\mathbf{x}_i) \frac{\partial}{\partial \phi} \left[ \ln q_{\phi}(\mathbf{x}_i) \right]\right/\sum_{\mathbf{x}\sim q_{\phi'}} w(\mathbf{x}_i),
\label{eqn.L_gradient}
\end{align}
where $w(\mathbf{x}_i) = \hat{p}(\mathbf{x})/q_{\phi'}(\mathbf{x})$, and $\hat{p}(\mathbf{x}) \propto {p}(\mathbf{x})$ is the nonnormalized target distribution. Notably, the samples $\left\{\mathbf{x}_i \right\}$ are drawn from $q_{\phi'}$, which is an NF model that is distinct from the one being trained, and the only calculation of $w(\mathbf{x}_i)$ involves the calculation of the likelihood. Therefore, the same data batch $\left\{\mathbf{x}_i \right\}$ and their associated weights $\left\{w(\mathbf{x}_i )\right\}$ can be used for multiple gradient descent steps without the need to reevaluate the likelihood of each sample, significantly reducing computational overhead. Our training procedure is summarized in Algorithm \ref{alg:train_with_Forward-KL}.

\begin{algorithm}[H]
\caption{Training normalizing flows with forward KL divergence}
\label{alg:train_with_Forward-KL}
\begin{algorithmic}[1]
\Require Learnable transformation $f_\phi$, nonnormalized  target distribution $\hat{p}(\mathbf{x})$, batch size $M$.
\State Sample a batch $\{\mathbf{z}_i\}_{i=1}^M$ from the latent distribution $p_z(\mathbf{z}) \sim \mathcal{N}(0, \mathbb{I})$.
\State Transform the data $\{\mathbf{z}_j\}_{j=1}^M$ to the target space to obtain $\mathbf{x}_j$: $\mathbf{x}_j$ $\gets$ $f_{\phi}(\mathbf{z}_i)$
\State For each sample in the batch, calculate  $q_{\phi}(\mathbf{x}_i)$ $\gets$ $  \left| \det \left( \left.d f_\phi^{-1}\right/d \mathbf{x} \right) \right| p_z(\mathbf{z}_i)$
\State Compute weights $w(\mathbf{x}_i) = \hat{p}(\mathbf{x}_i)/ q_\phi(\mathbf{x}_i)$.  \Comment{likelihood evaluation}
\State Minimize Forward KL loss: $\min_\phi\left[ -\left.\sum_{i=1}^M w(\mathbf{x}_i) \ln q_\phi(\mathbf{x}_i)\right/\sum_{i=1}^M w(\mathbf{x}_i)\right]$.    \Comment{train the network}
\State Repeat steps 1-5 until convergence.
\end{algorithmic}
\end{algorithm}

\subsection{Stabilizing the learning process with annealing }
\begin{figure}
\centering
\includegraphics[width=.7\linewidth]{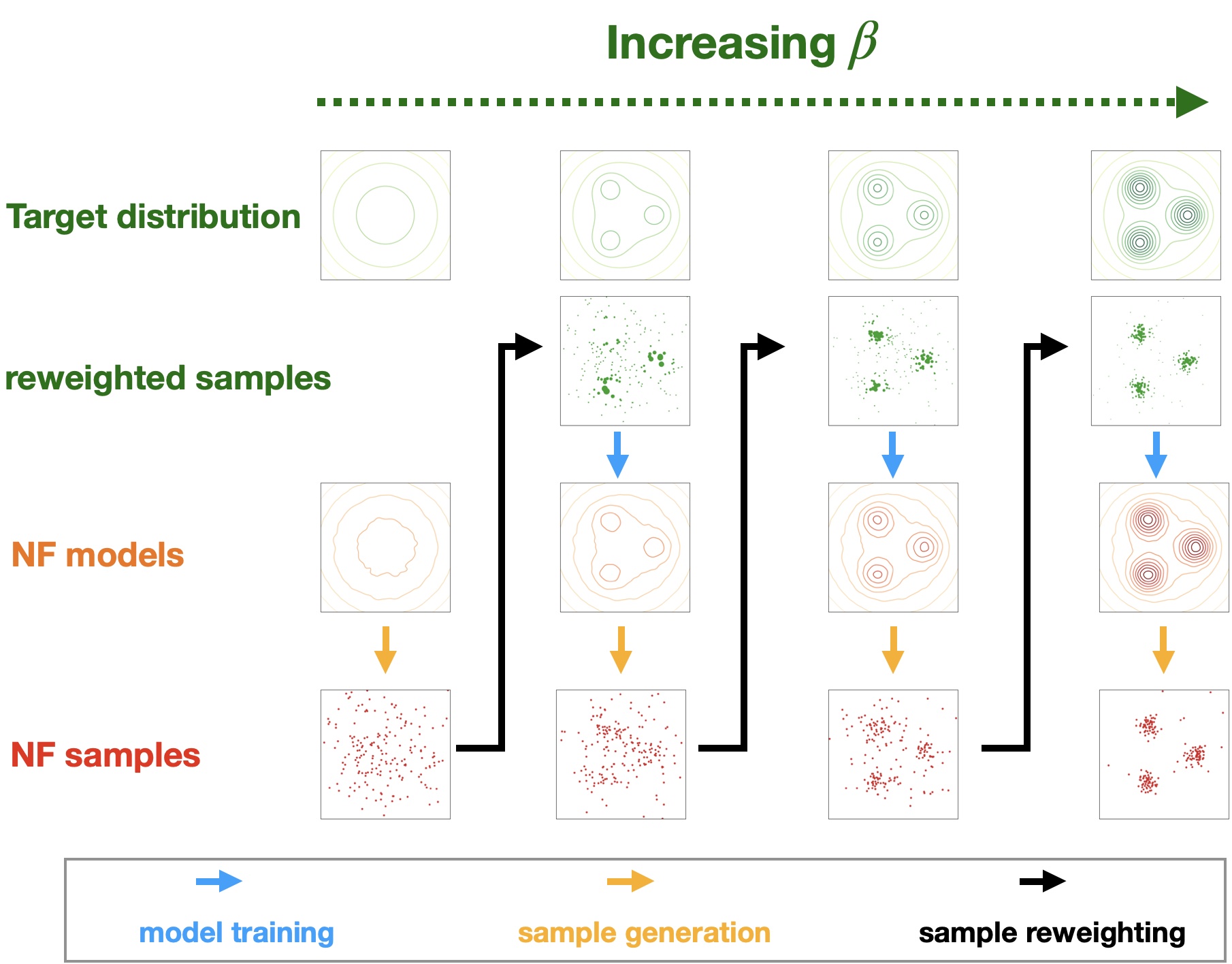}
\caption{\textbf{The annealing-sampling scheme.} 
The parameter $\beta$ increases to interpolate from a simple distribution that is close to the base distribution (top left) to the ultimate target distribution (top right).  At each value of $\beta$, samples generated from previously trained NFs are reweighted and used to train a new NF, from which samples are then drawn.  }
\label{fig:learning_scheme}
\end{figure}

For the numerical example that we consider below, we show that the procedure above yields different results each time that we train the NF independently, and, as mentioned previously, introducing training with the reverse loss and/or MCMC did not significantly improve the results.  
When the probability density of the NF model overlaps poorly with the target distribution, as is always the case initially, estimates of the weights of the samples under the target distribution and, in turn, averages, including the gradient of the loss function in \eqref{eqn.L_gradient}, tend to be inaccurate, which makes the training inefficient.

To address this issue, we introduce an annealing scheme in which the target distribution is gradually updated from a distribution that overlaps the base distribution well to the desired target distribution \cite{Simulated_Annealing}. 
Let $p_b(\mathbf{x})$ denote the base distribution and $p_u(\mathbf{x})$ denote an update distribution. We define the intermediate target distribution as $\hat{p}_{\beta}(\mathbf{x}) = p_b(\mathbf{x})  p_u(\mathbf{x})^\beta$, where $\beta\in[0,1]$ is the annealing parameter. Here, $\hat{p}_{\beta}$ is nonnormalized. The corresponding normalized distribution $p_{\beta} (\mathbf{x})  \propto \hat{p}_{\beta}(\mathbf{x})  $ smoothly transitions from the base distribution ($\beta=0$) to the full target distribution ($\beta=1$) as $\beta$ increases (Fig.\ \ref{fig:learning_scheme}).
In the numerical example that we consider below, the desired full target distribution is the posterior distribution for parameter estimation within a Bayesian framework. We identify the base distribution with the prior distribution and the update distribution with the likelihood function, so that the posterior is proportional to their product.

The success of annealing depends on its schedule.  We want the annealing to be sufficiently slow that the sampling is likely to converge to the desired precision but sufficiently fast that we limit unnecessary computation. Because we expect the schedule that balances reliability and efficiency to be system specific, we introduce an algorithm to determine the schedule based on the sampling. 
The algorithm is based on the effective sample size (ESS) \cite{liu1996metropolized}:
\begin{equation}
    n_{\mathrm{eff}}(\beta)=\frac{\left[\sum_i^n w\left(\mathbf{x}_i; \beta\right)\right]^2}{\sum_i^n\left[w\left(\mathbf{x}_i;\beta\right)\right]^2},
\label{eq.ess_NF}
\end{equation}
where $w(\mathbf{x}_i; \beta) = \hat{p}_{\beta}(\mathbf{x}_i)/q_{\phi}(\mathbf{x}_i)$ is the ratio between the nonnormalized target distribution $\hat{p}_{\beta}(\mathbf{x}_i)$ and the learned distribution $q_\phi(\mathbf{x}_i)$, as defined above.
If the density learned by the NF exactly matches the target distribution, the effective sample size is the actual sample size. In contrast, if the weight of one sample is much larger than the others, $n_{\mathrm{eff}}\approx1$. More generally, $n_{\mathrm{eff}}$ is between these limits.
During training with annealing, each increase in $\beta$ tends to push down $n_\mathrm{eff}$, and it recovers as the match between  $\hat{p}$ and $q_\phi$ improves.  We aim to keep $n_\mathrm{eff}$ near or above a threshold.





In practice, $n_\mathrm{eff}$ fluctuates significantly during training.  We thus base our algorithm on the exponential moving average:  \begin{equation}
    \bar{n}_\mathrm{eff} \leftarrow \lambda n_\mathrm{eff} + (1-\lambda) \bar{n}_\mathrm{eff},
\end{equation}
where $0\leq\lambda\leq 1$ controls the rate at which earlier contributions decay. Here we use $\lambda = 0.01$.

When $\bar{n}_\mathrm{eff}$ goes from below to above a fixed threshold $n^{*}$, we increase $\beta$.  Denoting the current and new $\beta$ values by $\beta_s$ and $\beta_{s+1}$, respectively,
we use the existing samples to solve numerically for the value of $\beta_{s+1}$ that satisfies $\gamma = n_{\mathrm{eff}}(\beta_{s+1})/n_{\mathrm{eff}}(\beta_{s})$, where $\gamma$ is a hyperparameter. In our study, we set $\gamma = 0.95$. 
The algorithm for updating the value of $\beta_s$ is summarized in Algorithm \ref{alg:beta_update}.

\begin{algorithm}[H]
\caption{$\beta_s$ update}
\label{alg:beta_update}
\begin{algorithmic}[1]
\Require Samples from NF model $\{\mathbf{x}_i\}_{i=1}^{n}$, annealing parameter $\beta_s$, discount factor of ESS $\gamma$, update distribution $p_u$, and base distribution $p_b$.
\Function{$\beta_{\text{update}}$}{$\{\mathbf{x}_i\}_{i=1}^n$, $\beta_s$, $\gamma$, $p_u$, $p_b$}
    \State Compute target distribution: $\hat{p}_{\beta_s}(\mathbf{x}) \gets p_b(\mathbf{x}) p_u(\mathbf{x})^{\beta_s}$
    \State Compute importance weights: $w(\mathbf{x}_i; \beta_s) \gets \hat{p}_{\beta_s}(\mathbf{x}_i)/q_{\phi}(\mathbf{x}_i)$ for $i=1$ to $n$
    \State Compute ESS: $n_{\text{eff}}(\beta_s) \gets \left(\sum_{i=1}^n w_i\right)^2 / \sum_{i=1}^n w_i^2$
    \State Solve $\beta_{s+1}$ such that $n_{\text{eff}}(\beta_{s+1}) = \gamma \cdot n_{\text{eff}}(\beta_s)$ \Comment{solve numerically with root-finding algorithm}
    \State \Return $\beta_{s+1}$
\EndFunction
\end{algorithmic}
\end{algorithm}

\subsection{Mixing samples}

The annealing procedure above generates data (and a trained model) at each value of $\beta_s$.  We can use all these data, rather than just the data sampled from the current NF, by forming a mixture model $q_m$.
Following \cite{mbar_reweighting}, we can think of the $N_k$ samples $\{\mathbf{x}_{n,k}\}_{n=1}^{N_k}$ drawn from each of the $K$ NF models $\{q_{\phi_k}\}_{k=1}^K$ as being drawn randomly from 
\begin{equation}
\label{eq.mixed_distribution}
    {q_m\left(\mathbf{x}\right)} =\frac{\sum_k^K N_k  q_{\phi_k}\left(\mathbf{x}\right) }{\sum_k^K N_k }.
\end{equation}
That is, the probability of a sample $\mathbf{x}$ is the  weighted average of its probabilities under the $K$ NF models. 
%
Then, when training using Algorithm \ref{alg:train_with_Forward-KL}, we compute the weights $w(\mathbf{x})$ as
\begin{equation}
\label{eq.mbar_weights}
    w (\mathbf{x}; \beta_s) = \frac{\hat{p}_{\beta_s}\left(\mathbf{x}\right)}{q_m\left(\mathbf{x}\right)}.
\end{equation}
We note that, because each $q_{\phi}$ is normalized by construction, $q_m$ is normalized as well. We summarize our procedure for training with annealing in Algorithm \ref{alg:full_algorithm}.

\begin{algorithm}[H]
\caption{Annealing protocol for training Normalizing Flows}
\label{alg:full_algorithm}
\begin{algorithmic}[1]
\Require Learnable NF model $q_{\phi_k}$, 
update distribution $p_u$, and base distribution $p_b$, batch size $B$, network update steps $J$, maximum number of batches $M$, decay rate of exponential moving average $\lambda$ 

\State Initialize $\beta_s \gets 0$, $\bar{n}_{\text{eff}} \gets 0$, $k \gets 1$
 \State Define initial target distribution:  $\hat{p}_{\beta_s}(\mathbf{x}) \gets p_p(\mathbf{x})$
\While{$\beta_s<1$}
    \State sample a batch of data $\{\mathbf{x}_{n,k}\}_{n=1}^B$ from $q_{\phi_k}(\mathbf{x})$ 
    \State compute the likelihood and prior for the samples $\{\mathbf{x}_{n,k}\}_{n=1}^B$
    \State estimate the effective sample size $n_{\text{eff}}$
    \State update the exponential moving average of effective sample size $\bar{n}_{\text{eff}}$ $\gets$ $\lambda \cdot \bar{n}_{\text{eff}} + (1-\lambda) n_{\text{eff}} $
     \If{ ($\bar{n}_{\text{eff}}> n^*)$}
        \State Update $\beta_s$ according to Algorithm \ref{alg:beta_update}: $\beta_s \gets  \beta_{\text{update}}\left( \{\mathbf{x}_{n,k}\}_{n=1}^B, \beta_s, \gamma, p_u, p_b\right)$ 
        \State Update the target distribution: $\hat{p}_{\beta_s} \gets p_b(\mathbf{x}) p_u(\mathbf{x})^{\beta_s} $
    \EndIf
\State Add $\{\mathbf{x}_{n,k}\}_{n=1}^B$ to the training dataset:
\Statex \hspace{2em} 
 $\bigcup_{j=\max(1, k-M)}^k\{\mathbf{x}_{n,j}\}_{n=1}^{B}\gets \left\{\bigcup_{j=\max(1, k-M)}^{k-1}\{\mathbf{x}_{n,j}\}_{n=1}^{B}, \{\mathbf{x}_{n,k}\}_{n=1}^B\right\}$
\State Update the sample weights according to  \eqref{eq.mixed_distribution} and \eqref{eq.mbar_weights}
\State Minimize the forward KL divergence using a mini-batch of size $B$, randomly sampled from the dataset: \Statex \hspace{2em} 
 $\min_{\phi_k} \left.\sum_{i=1}^B w (\mathbf{x}_i; \beta_s) \log q_{\phi_k}(x_i)\right/\sum_{i=1}^B w (\mathbf{x}; \beta_s)$ for $J$ steps.
\State $k \gets k+1$ 
\EndWhile
\end{algorithmic}
\end{algorithm}

\section{Bayesian parameter estimation}

The numerical example that we present involves fitting time-series data to estimate parameters of an ODE model. The fitting is guided by a Bayesian framework, which we review in Section \ref{sec:BayesBackground}. We discuss how specific quantities can be computed using NFs that are trained with annealing in Section \ref{sec:NFmarginals}.

\subsection{Background}
\label{sec:BayesBackground}


When fitting data, we seek to determine the probability of the parameters, $\boldsymbol{\theta}$, given the data, $D$, and a model, $M$---i.e., $P(\boldsymbol{\theta}\mid D, M)$.
Given our prior assumptions about the distribution of the parameters, $P(\boldsymbol{\theta} \mid M)$, and the likelihood of observing the data given the parameters, $P(D\mid \boldsymbol{\theta}, M)$, we can calculate the posterior distribution $P(\boldsymbol{\theta}\mid D, M)$  by Bayes' theorem:
\begin{equation}\label{eq:bayes}
    P(\boldsymbol{\theta} \mid D, M) = \frac{P(D\mid \boldsymbol{\theta}, M)P(\boldsymbol{\theta}, M)}{P(D\mid M)}. 
\end{equation}
%
The factor $P(D\mid M) \equiv \int P(D\mid \boldsymbol{\theta}, M)P(\boldsymbol{\theta}\mid M) d\boldsymbol{\theta}$ in the denominator of \eqref{eq:bayes} is known as the marginal likelihood or the model evidence.  It can be used to compare models $M_1$ and $M_2$, again through Bayes' theorem:
\begin{equation}\label{eq:evidence}
    \frac{P(M_1\mid D)}{P(M_2\mid D)} = \frac{P(D \mid M_1)P(M_1)}{P(D \mid M_2)P(M_2)}.
\end{equation}
When our prior beliefs in models $M_1$ and $M_2$ are equal, $P(M_1)/P(M_2)=1$, and the right hand side of \eqref{eq:evidence} reduces to the ratio of the marginal likelihoods, known as the Bayes factor.

\subsection{Estimating marginal likelihoods}
\label{sec:NFmarginals}


As shown above, marginal likelihoods are key to comparing models.  We consider two ways to compute marginal likelihoods when training NFs with annealing.
The first way is by importance sampling from the learned posterior distribution $q_\phi$ for $\beta=1$:
\begin{equation}
P(D \mid M)
= \left< 
 \frac{P(D \mid \boldsymbol{\theta},M)P(\boldsymbol{\theta} \mid M)}{q_{\phi}\left(\boldsymbol{\theta}\right)} 
\right>_{q_{\phi}\left(\boldsymbol{\theta}\right)},
\label{eq.Bayes_factor_IS}
\end{equation}
where now $\mathbf{x} \equiv\boldsymbol{\theta}$, and we  write \eqref{eq.Bayes_factor_IS} as an average by inserting $q_\phi/q_\phi=1$ into the definition of the marginal likelihood and then using the fact that $q_\phi$ is normalized.
The second way is to use thermodynamic integration (TI) \cite{Bayes_factor_TI} to combine the data obtained during annealing:
\begin{equation}
\ln P(D \mid M)=\int_0^1 \langle\ln P(D \mid  \boldsymbol{\theta}, M)\rangle_{P_\beta(\boldsymbol{\theta})} d \beta,
\label{eq.TI}
\end{equation}
where the expectations $\langle \cdot \rangle_{p_\beta}$ are calculated by reweighting the samples drawn from the NF models $q_\phi$ trained at each $\beta$ to the target distribution $P_\beta(\boldsymbol{\theta}) \propto P(D\mid \boldsymbol{\theta},M)^{\beta}P(\boldsymbol{\theta}, M) $ (see Supplementary Materials, Section \ref{sec.TI}). 
We compare these two approaches for our numerical example.\\

\section{\large{Numerical example}}

As mentioned previously, we test our annealing scheme by fitting time-series data to estimate parameters for a set of ODEs.  
As we show, there are strong correlations between the parameters, their distribution is multimodal, and the likelihood is computationally costly to evaluate.  These features make this application a challenging test. 


\subsection{Repressilator model}

\begin{figure}[hbt]
\centering
\includegraphics[width=0.8\linewidth]{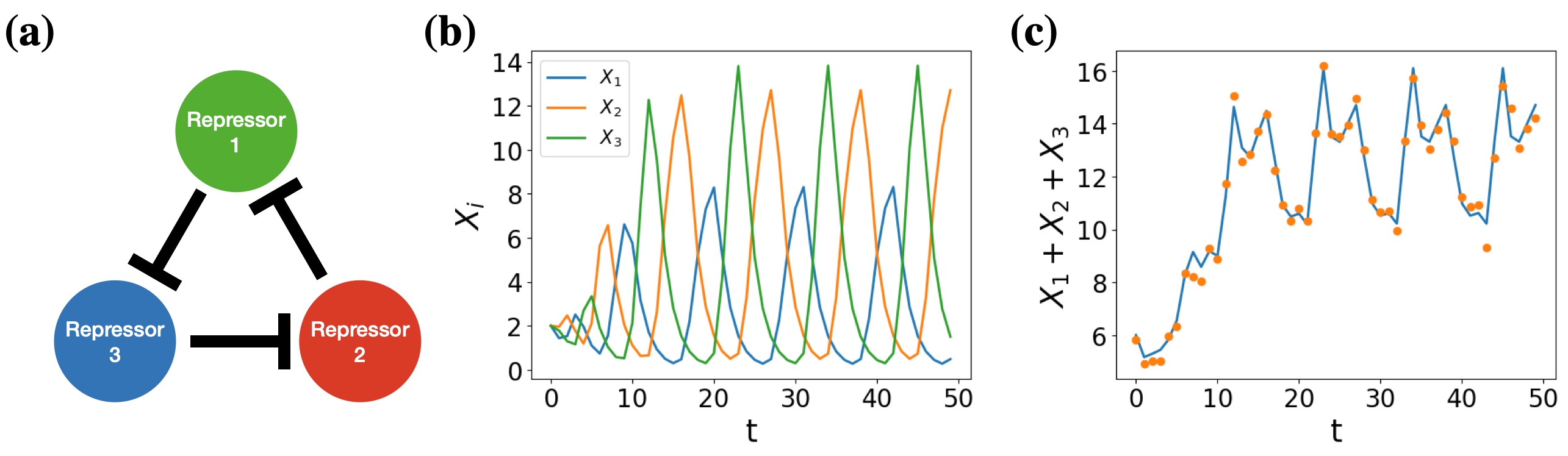}
\caption{\textbf{Repressilator model.} (a) Schematic of the system; each circle represents a gene product, and $i\dashv j$ represents repression of $j$ by $i$. (b) Solution used to generate the data for fitting; the parameter values are $X_i(t_0) = 2$ for all $i$, $\alpha_1 = 10$, $\alpha_2 = 15$, $\alpha_3 = 20$, $m=4$, and $\eta = 1$.  (c) Time series of the total concentration of gene products (blue line) and the simulated observable (orange dots) produced by adding Gaussian noise with variance 0.25.} 
\label{fig:repressilator}
\end{figure}

The specific set of ODEs that we study represents a model of the repressilator, a biochemical oscillator that comprises a cycle of three gene products that each represses expression of another  (Fig.\ \ref{fig:repressilator}) \cite{repressilator,biocircuits}:
\begin{equation}
\frac{d X_i}{d t}=\frac{\alpha_i}{1+X_{p(i)}^m}-\eta X_i\quad\text{for}\quad \quad i=1,2,3
\label{eq.repressilator}
\end{equation}
where $X_i$ represents the concentrations of gene product $i$, $\alpha_i$ is its production rate, $m$ is a Hill coefficient, $\eta$ is the degradation rate, and $p(i)\equiv (i\mod 3)+1$ is a periodic indexing function. We assume that $m$ and $\eta$ are the same for all gene products. 

We generate data by integrating the ODEs with the explicit fifth-order Runge-Kutta method of Tsitouras \cite{TSI5} from $t_0 = 0$ to $t_1 =30.0$, saving states with a time interval of 0.6. The  parameter values are $X_i(t_0) = 2$ for all $i$, $\alpha_1 = 10$, $\alpha_2 = 15$, $\alpha_2 = 20$, $m=4$, and $\eta = 1$. We treat the total concentration of gene products (i.e., $X_1+X_2+X_3$) as the observable and add Gaussian noise with a variance of $\sigma^2 =0.25$ to mimic experimental uncertainty. The time series of the concentration of each gene product and the observable are shown in Figs. \ref{fig:repressilator}b and \ref{fig:repressilator}c. 

Including the initial concentrations of the gene products, there are eight parameters to estimate: $\boldsymbol{\theta} = \{X_1(t_0), X_2(t_0), X_3(t_0), \alpha_1, \alpha_2, \alpha_3, m, \eta \}$. Let $D(t)$ denote the measurement data at time $t$, and $\mathbf{X}(t, \boldsymbol{\theta})$ be the solution of \eqref{eq.repressilator} for a given $\boldsymbol{\theta}$ at time $t$. The function $\hat{D}(\mathbf{X}(t, \boldsymbol{\theta}))$ maps from the model variables $\mathbf{X}(t)$ to the observables. Assuming a Gaussian measurement noise model with variance $\sigma^2$, the likelihood of observing the data $D = \{D(t)\}_{t=1}^T$ is:
\begin{equation}
    p(D \mid \boldsymbol{\theta}, M) = \prod_{t=1}^T \frac{1}{\sqrt{2\pi\sigma^2}} \exp\left(-\frac{\|D( \mathbf{X}(t)) - \hat{D}(\mathbf{X}(t;\boldsymbol{\theta}))\|^2}{2\sigma^2}\right).
\end{equation}
In cases where the ODE solver fails to provide a solution, we assign  $\hat{D}_i(\boldsymbol{\theta}) = 200$ for all time points. We define the prior distribution as
\begin{equation}
p(\boldsymbol{\theta} \mid M) \propto \exp\left[ -\frac{1}{2} (\boldsymbol{\theta} - \boldsymbol{\mu})^T \boldsymbol{\Sigma}^{-1} (\boldsymbol{\theta} - \boldsymbol{\mu}) \right],
\end{equation}
where
$\boldsymbol{\mu}^T = 
(2, 2, 2, 15, 15, 15, 5, 5)$
and $\boldsymbol{\Sigma} = 
\mathrm{diag}(4, 4, 4, 25, 25, 25, 25, 25)$.
Due to the symmetry of the ODEs and that the observable does not resolve individual species, the posterior has three modes. 

\subsection{NF network architecture and training}

We train an NF for each value of the annealing parameter $\beta$.
At the beginning of the annealing process, the target distribution with $\beta_s=0$ corresponds to the prior, which we take to be a simple, unimodal, and easy-to-evaluate distribution. Consequently, the network can easily fit this distribution, even with randomly initialized parameters. For subsequent training, as $\beta$ increases, the target distribution becomes more complex. Rather than retraining from scratch, we use the previously trained model as a start.

For the NFs, we use multilayer perceptrons (MLPs) with three hidden layers for both the scaling functions $a^\ell_\phi$ and the translation functions $b^\ell_\phi$. Each hidden layer is fully connected, and the layer size is three times the input dimension, which is half the number of parameters (so that the total across $a_\phi$ and $b_\phi$ is equal to the number of parameters). The hidden layers utilize the Gaussian Error Linear Unit (GELU) activation function, and the output layer uses a linear transformation to reduce the output dimension to half the number of parameters.

We initialize the weights randomly and train the NF with the Adam optimizer \cite{adam} with a learning rate of $10^{-4}$, first moment decay rate of $0.9$, and a second moment decay rate of $0.99$.  We clip the norm of the gradient to $10^6$ to stabilize the training process \cite{gradient_clip}.

\subsection{Stable and efficient sampling with an NF-based sampling scheme}

\begin{figure}
\centering
\includegraphics[width=0.8\linewidth]{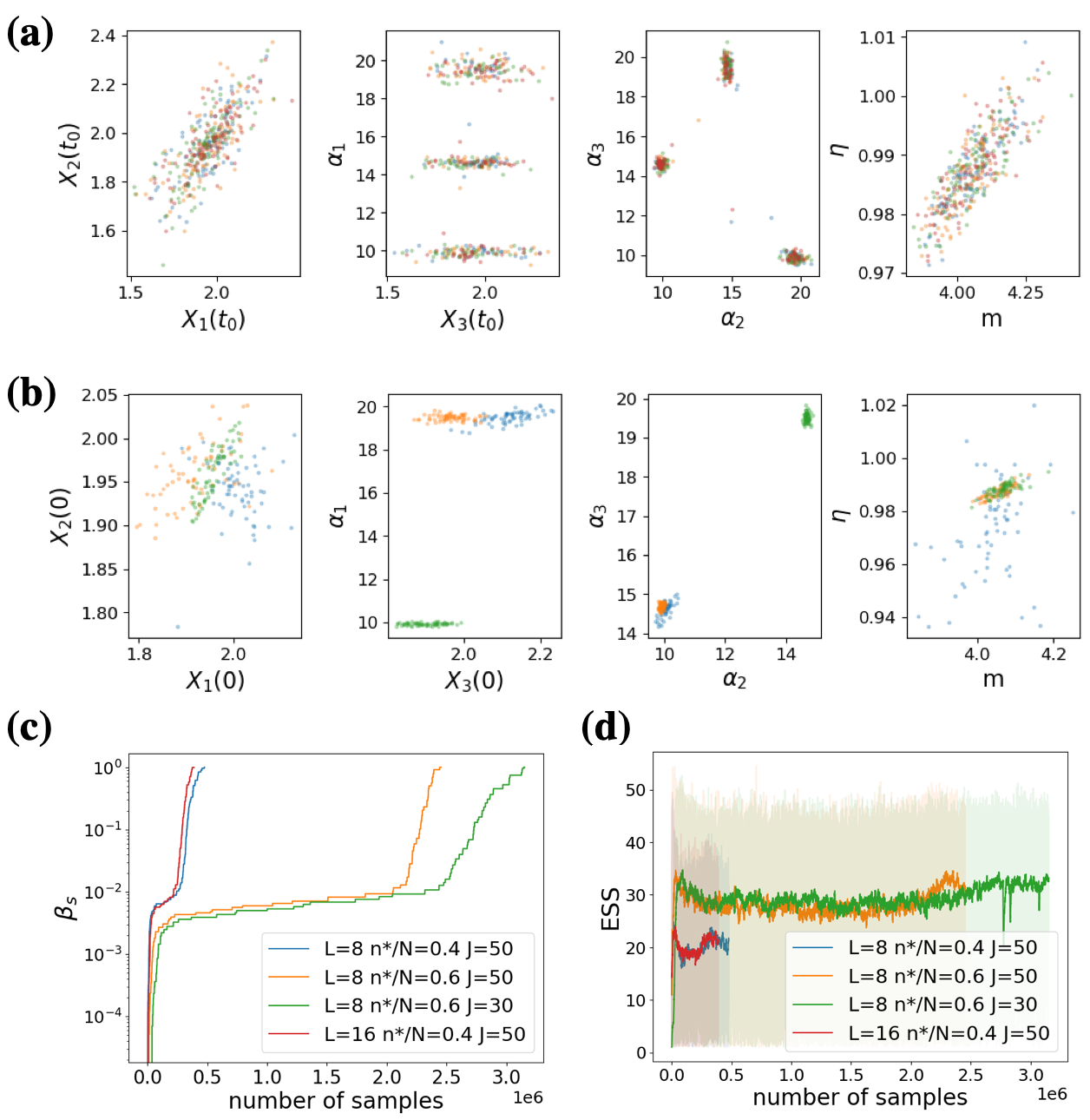}  
\caption{\textbf{Annealing with a schedule based on the effective sample size (ESS) mitigates mode collapse.}
(a) Samples from NF models trained with the annealing protocol. Colors correspond to different hyperparameter choices, as specified in (c) and (d). 
(b) Samples from NF models trained with fixed $\beta=1$. Different colors distinguish samples from three models trained with distinct initial weights and random seeds. Note that the scales in (a) and (b) are different.
(c) The change of $\beta_s$ in training from three independent runs with different parameters.
(d) The effective sample size (ESS). Translucent lines show the instantaneous ESS values ($n_\mathrm{eff}$) and opaque lines show their exponetial moving average ($\bar{n}_\mathrm{eff}$) with $\lambda= 0.01$. }
\label{fig:NF_samples}
\end{figure}

We apply the NF sampling scheme with various hyperparameter choices to estimate parameters for the repressilator model. Specifically, we vary the number of layers in RealNVP ($L$), the ESS threshold ($n^*$), and the number of steps for network updates before updating the training dataset ($J$ in Algorithm \ref{alg:full_algorithm}).
As shown in Fig.\ \ref{fig:NF_samples}a, the NFs successfully capture the three modes of the system. Throughout the training process, we employ Algorithm \ref{alg:beta_update} to update the $\beta_s$ values automatically, and the rate of change varies considerably, with all runs exhibiting a slowdown at $\beta_s \approx 0.06$ (Fig.\ \ref{fig:NF_samples}c). The  ESSs remain relatively stable throughout training (Fig.\ \ref{fig:NF_samples}d), consistent with the observation above that the NFs avoid mode collapse.    
In contrast, NFs fail to reliably sample all three modes and often become stuck in local minima when they are trained with a fixed $\beta_s$ value (Fig.\ \ref{fig:NF_samples}b) or with 
the schedule employed by Friel and Pettitt \cite{friel2008powerPosteriors}: $\beta_s =( s/1000 )^4$ for $s = 1, 2, \cdots, 1000$ 
(Supplementary Fig.\ \ref{fig:preset_schedule}).

For the hyperparameter combinations that we test, we find that the ESS threshold $n^*$ affects the efficiency of the training to the greatest degree. When the threshold is higher, $\beta_s$ increases more slowly.
When the ESS threshold $n^*$ is too small, intermediate target distributions can be insufficiently sampled, and features of the distribution that are important to learn are missed (Supplementary Fig.\ \ref{fig:samples_ESS-threshold-0.2}). Conversely, if $n^*$ is too large, the network can fail to reach to reach the threshold, and $\beta_s$ can become stuck (see Supplementary Fig.\ \ref{fig:compare_ESS_threshold}). Therefore, in applications, a tradeoff between accuracy and efficiency needs to be considered. 
For the two runs with $n^*=0.4$, the run with $L=16$ requires significantly more computation time at each $\beta_s$ and almost the same number of samples as the run with $L=8$, making the $L=16$ run more computationally costly overall (Figs.\ \ref{fig:NF_samples}c and \ref{fig:NF_samples}d and Table \ref{Table.Bayes_factor}). 
By contrast, we found that the choice of $J$, the number of steps to update the NF model prior to updating the samples, has little effect on training efficiency (Figs.\ \ref{fig:NF_samples}c and \ref{fig:NF_samples}d and Table \ref{Table.Bayes_factor}). 
Among runs that gave comparable results, that with $L=8$, $n^*/N = 0.4$, and $J=50$ required the smallest computation time by far.

\begin{table}[bt]
\begin{center}
\begin{tabular}{lccccc}
\hline
Sampling method                          & NF & NF & NF & NF   & MCMC \\ \hline\hline
Number of layers ($L$)             & 8                                 & 8                                & 8                                    & 16                                & --                                  \\ \hline
ESS ratio threshold ($n^*/N$)          & 0.4                               & 0.6                              & 0.6                                  & 0.4                               & --                                  \\ \hline
Number of NN updates between samples ($J$)             & 50                                & 50                               & 30
      & 50                                & --                                  \\ \hline
Total number of samples & $4.7 \times 10^5$            & $2.5\times10^6 $                 & $3.2\times10^6 $
             & $\mathbf{3.9}\times \mathbf{10^5}$             & $2.5\times10^7$          \\ \hline
Computation time & \textbf{5.2 hrs}               & 26.7 hrs                         & 31.6 hrs                                & 25.8 hrs                       & 58 hrs                             \\ \hline
$\log P(D|M)$ ($\beta_s=1$)    & --35.61                  & --35.59                    & --35.58                           & --35.58                  &                              \\ \hline
$\log P(D|M)$ (TI)     & --35.90                  & --35.77                    & --35.80                           & --35.55                  & $-35.9 \pm 0.2$                           \\ \hline
\end{tabular}
\end{center}
\caption{\textbf{Performance comparison.} The computation times were recorded on an Intel Xeon Gold 6248R CPU. The marginal likelihoods reported for importance sampling with $\beta_s=1$ exclude samples with large weights, as discussed in the text. To estimate the variance of the estimate from MCMC, we conducted four independent sampling runs with different random seeds. }
\label{Table.Bayes_factor}
\end{table}

\subsection{Comparison with MCMC}

We compare the NF sampling with MCMC sampling. We employ an MCMC approach which evolves an ensemble of parameter sets (walkers) and uses the variation between them to inform proposals (moves) \cite{goodman2010ensemble,foreman2013emcee}. Specifically, with equal probabilities, we attempt stretch moves  \cite{goodman2010ensemble,foreman2013emcee} and differential evolution moves \cite{braak2006markov,DEMove}, both of which translate walkers in the direction of vectors connecting two walkers (in the former, the walker one attempts to move is one of the two walkers defining the vector, and, in the latter, it is not). 
We use a scale parameter of $2$ for the stretch moves and  $0.595$ for the differential evolution moves; the latter is based on the formula $2.38/\sqrt{2d}$, where $d$ is the dimension of the parameter space \cite{braak2006markov,DEMove} (here, $d=8$).  For the differential evolution moves, each parameter value was additionally perturbed by a random amount drawn from a Gaussian distribution with variance $10^{-5}$. 

As for the NF sampling, we find it necessary to anneal the MCMC sampling. Moves are thus accepted according to the Metropolis-Hastings criterion \cite{metropolis1953equation} based on 
$p_{\beta_s}(\boldsymbol{\theta}| D, M) = p(\boldsymbol{\theta}|M) p(D|\boldsymbol{\theta}, M)^{\beta_s}$. The schedule of $\beta_s$ follows the scheme employed by Friel and Pettitt \cite{friel2008powerPosteriors}, with $\beta_s =( s/1000 )^4$ for $s = 1, 2, \cdots, 1000$.  The initial walkers are directly sampled from the prior. We attempt 1500 moves for each $\beta_s$; we found that runs with $1000$ moves for each $\beta_s$ were more prone to generate outliers from the target distribution (about 1/3 of runs with 1000 moves have outliers, compared with about 1/5 of runs with 1500 moves; examples of runs with outliers are shown in Supplementary Fig.\ \ref{fig:MCMC_outliers}). 
Given the samples collected at each $\beta_s$, we estimate the marginal likelihood using the thermodynamic integration formula in \eqref{eq.TI}.

 
With this annealing schedule, the ensemble MCMC also captures the three maxima of the system. However, the MCMC requires one to two orders of magnitude more samples than the NF schemes, depending on the choice of hyperparameters (Table\ \ref{Table.Bayes_factor}). The effective sample sizes and acceptance rates for the MCMC simulations are provided in the Supplementary Figs.\ \ref{fig:MCMC_ESS} and \ref{fig:acceptance_rates}. Even accounting for the computational time required to train the networks, the NF-based sampling scheme achieves around a ten-fold speedup compared to MCMC sampling. 

\subsection{Reliable estimates of marginal likelihoods}

\begin{figure}[bt]
\centering
\includegraphics[width=0.8\linewidth]{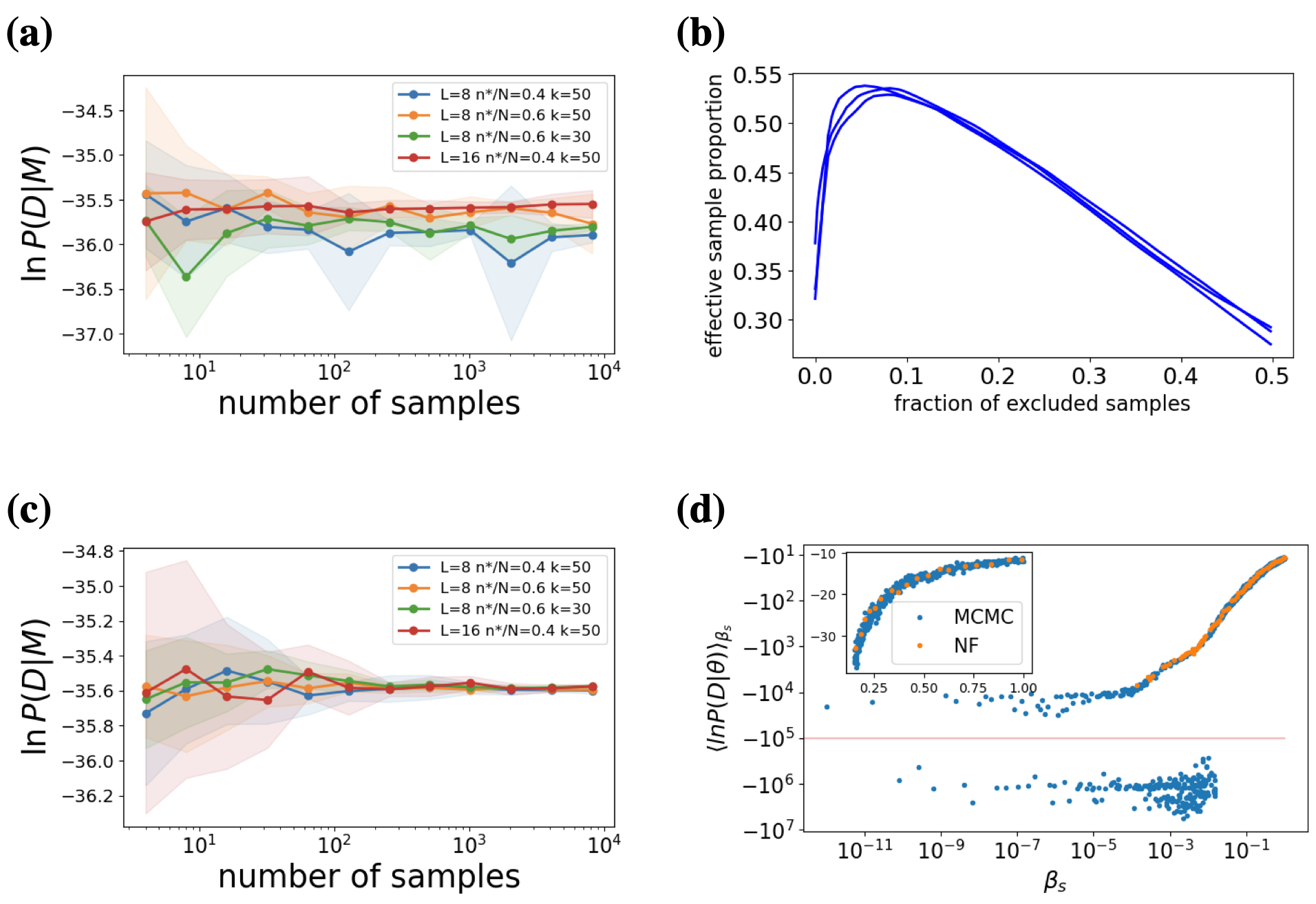} %
\caption{{\bf Estimating marginal likelihoods.} 
(a, c) Marginal likelihood estimates for different sample sizes. Shaded regions indicate standard deviations from 10 independent sets of samples. In (a), all samples are used for the estimates. In (c), samples are excluded to maximize the effective sample size. (b) Variation of the effective sample size (ESS) as samples with large weights are excluded. Results for three independent batches of samples are shown. (d) Integrand for thermodynamic integration (TI). The red solid line indicates the cutoff below which data points are excluded. The inset provides an expanded view for $\beta_s>0.2$.   }
\label{fig:marginal_likelihood}
\end{figure}



In this section, we examine the performance of the NF and MCMC sampling schemes in terms of the estimates that they yield for marginal likelihoods.  In Fig.\ \ref{fig:marginal_likelihood}a, we show the marginal likelihood computed using the importance sampling formula in \eqref{eq.Bayes_factor_IS} for NF sampling with the four hyperparameter combinations considered previously. We observe that, not only do the different hyperparameter combinations give different results, but the variance across independent runs for a given hyperparameter combination does not decrease as the number of samples increases.
We interpret this behavior to result from the NF inaccurately accounting for tail probabilities, such that occasional samples give disproportionately large contributions to \eqref{eq.Bayes_factor_IS}.
%
To address this issue, we rank order the samples by their weights and exclude those samples with the highest weights until the effective sample size of the remaining samples is maximized (Fig.\ \ref{fig:marginal_likelihood}b).  While in principle selectively excluding samples biases the results, as shown in Fig.\ \ref{fig:marginal_likelihood}c, it markedly improves the convergence of estimates of the marginal likelihood, both for a given hyperparameter combination but also across different hyperparameter combinations (see also Table\ \ref{Table.Bayes_factor}).  

We can compare the above estimates to those from the thermodynamic integration of the NF or MCMC samples.
As shown in Fig.\ \ref{fig:marginal_likelihood}d, the  integrand of \eqref{eq.TI} agrees well for the NF and MCMC samples at $\beta_s> 10^{-4}$.  However, it exhibits significant variance for the MCMC samples when $\beta_s < 0.016$. This is due to the existence of samples with ODE parameters which prevent the ODE from being numerically solved with the desired accuracy, leading to extremely small likelihoods. To address this issue, we exclude $\beta_s$ values with logarithms of the marginal likelihood less than $-10^5$.  Then we employ the trapezoidal rule \cite{Numerical_Recipes} to approximate the integral with the remaining points. We obtain good agreement between the NF with different hyperparameter combinations and the MCMC; overall, the estimates from thermodynamic integration are slightly larger in magnitude than those from importance sampling, perhaps reflecting the bias introduced by excluding samples with large weights in the case of importance sampling.\\


\section{Discussion}

In this study, we explored NFs for parameter estimation in a Bayesian framework, with a focus on mitigating mode collapse. Our main innovation is an adaptive annealing scheme based on the ESS. This adaptive scheme not only eliminates the need to choose a schedule but, more importantly, automates the allocation of computational resources across various annealing stages so as to capture multiple modes when used together with a mode-covering loss (here, the forward KL divergence). For the numerical example that we considered---estimating marginal likelihoods of a model of a biochemical oscillator---we achieved about a ten-fold savings in computation time relative to a widely-used MCMC approach. 

Our approach is general, and we expect it to be useful for other sampling problems, especially ones in which making MCMC proposals is computationally costly.
Furthermore, we note that the gradient of the likelihood is not needed, so the approach can be applied even when the likelihood (or energy function) has non-differentiable components. That said, the approach remains to be tested on problems with larger numbers of degrees of freedom. One potential issue is that importance sampling is a key element of the approach, and the variance of the weights in importance sampling can become large in high-dimensions \cite{au_beck_2003,KATAFYGIOTIS2008208}. 

These considerations point to elements of the approach that provide scope for further engineering.
While effective in the present case, alternatives to the ESS could be considered to quantify the similarity between the model and target distributions.  Similarly, we used RealNVP \cite{RealNVP} for the NF architecture, but other architectures \cite{glow,maskedautoregressiveflow,ffjord,neuralsplineflows} and transport schemes \cite{souveton2024fixed,stochasticinterpolants} could be considered. Finally, the adaptive annealing scheme could be combined with other approaches for mitigating mode collapse \cite{mode-collapse_NF,mate2023learning,vaitl2022gradients,midgley2023flow}.  We believe these directions merit investigating in the future.\\

\section*{Acknowledgments}
We thank Jonathan Weare for critical readings of the manuscript and Michael Rust and Yujia Liu for helpful discussions.  This work was supported in part by Chicago Center for Theoretical Chemistry and Eric and Wendy Schmidt AI in Science Postdoctoral Fellowships to YW and  by grants from the NSF (DMS-2235451) and Simons Foundation (MP-TMPS-00005320) to the NSF-Simons National Institute for Theory and Mathematics in Biology (NITMB).

\newpage 
{\Large\textbf{Supplementary Materials}}

\setcounter{section}{0}
\renewcommand{\thefigure}{S\arabic{figure}}
\setcounter{figure}{0} 


\section{Thermodynamic integration}
\label{sec.TI}
Here we derive the formula for thermodynamic integration \eqref{eq.TI}.  Given the unnormalized target distribution
     $\hat{P}_{\beta}(\theta) = P(D\mid \boldsymbol{\theta},M)^{\beta}P(\boldsymbol{\theta}, M)$,
we define the normalization constant (partition function) $Z_{\beta}$ and normalized distribution $P_{\beta}$:
\begin{equation}
\begin{aligned}
Z_{\beta} = \int \hat{P}_\beta(\theta) d \theta \quad\text{and}\quad
 P_{\beta}(\boldsymbol{\theta}) = \frac{\hat{P}_\beta(\boldsymbol{\theta})}{Z_\beta}
\end{aligned}
\end{equation}
We then differentiate $\ln Z_\beta$ with respect to $\beta$:
\begin{equation}
\begin{aligned}
\frac{\partial \ln Z_\beta}{\partial \beta} & 
=\frac{1}{Z_\beta} \frac{\partial Z_\beta}{\partial \beta} \\
& =\frac{1}{Z_\beta} \int \frac{\partial \hat{P}_\beta(\boldsymbol{\theta})}{\partial \beta} d \boldsymbol{\theta} \\
& =\int \frac{1}{\hat{P}_\beta(\boldsymbol{\theta})}\frac{
\partial \hat{P}_\beta(\theta)}{\partial \beta} \frac{\hat{P}_\beta(\boldsymbol{\theta})}{Z_\beta} d \theta \\
& =\int \frac{\partial \ln \hat{P}_\beta(\boldsymbol{\theta})}{\partial \beta} P_\beta(\boldsymbol{\theta}) d \theta \\
& =\left< \frac{\partial \ln \hat{P}_\beta(\boldsymbol{\theta})}{\partial \beta}\right>_{P_{\beta}} \\
& =  \left< \ln P(D\mid \boldsymbol{\theta},M) \right>_{P_{\beta}}.
\end{aligned}
\end{equation}
Integrating over $\beta$ yields
\begin{equation}
\ln P(D \mid M)= \ln Z_1 - \ln Z_0 = \int_0^1 \langle\ln P(D \mid  \boldsymbol{\theta}, M)\rangle_{P_\beta(\boldsymbol{\theta})} d \beta,
\end{equation}
where $\ln Z_0= \ln \int P(\boldsymbol{\theta}, M)=0$ because $P(\boldsymbol{\theta}, M)$ is normalized.

\section{Effective sample size and acceptance rates in MCMC}
The ESS of MCMC measures that number of independent samples effectively obtained from a correlated chain \cite{sokal1997monte,carlin2008bayesian}. For each parameter, we computed the ESS using the formula:
\begin{equation}
    n_\mathrm{eff}=\frac{N}{1+2 \sum_{\tau=1}^{\infty} \rho_k}.
\label{eq.mcmc_ess}
\end{equation}
Here $N$ is the total number of samples, and 
\begin{equation}
    \rho_\tau = \frac{\sum_{t=1}^{N-\tau}\left(\theta_t-\bar{\theta}\right)\left(\theta_{t+\tau}-\bar{\theta}\right)}{\sum_{t=1}^N\left(\theta_t-\bar{\theta}\right)^2}
\end{equation} 
represents the autocorrelation at lag time $\tau$ \cite{geyer1992practical}. Equation\ \eqref{eq.mcmc_ess} provides the univariate ESS for each variable. Examples are shown in Supplementary Fig.\ \ref{fig:MCMC_ESS}. Since the ESS for NF samples and MCMC samples are defined differently (compare \eqref{eq.ess_NF} and \eqref{eq.mcmc_ess}), a direct comparison between them does not accurately reflect the relative efficiency of these methods. However, it should be noted that the ESS for NF samples is consistently close to $1$, indicating that the samples generated by the NFs are representative of the target distribution. In contrast, ESS values for MCMC are much smaller than $1$, making clear the strong correlations between successive samples.\\

\begin{figure}[h]
\centering
\includegraphics[width=0.6\linewidth]{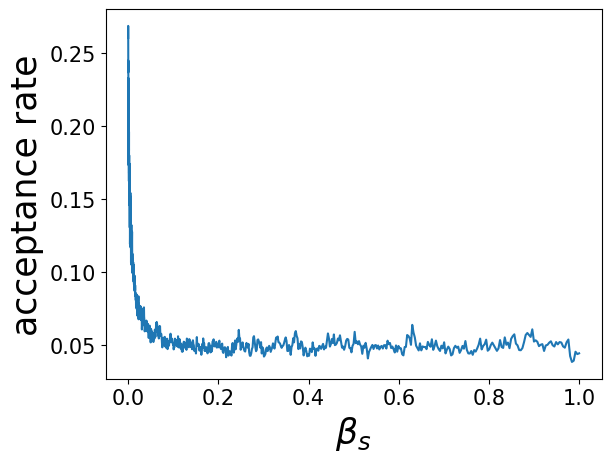} %
\caption{Acceptance rates in the MCMC simulations.}
\label{fig:acceptance_rates}
\end{figure}

We also monitor the change of acceptance rates in MCMC.  The acceptance rate is highest at small $\beta_s$ and decreases as $\beta_s$ increases, eventually stabilizing at a low value around $0.04$ (Supplementary Fig. \ref{fig:acceptance_rates}).

\begin{figure}[h] 
\centering
  \includegraphics[width=.8\linewidth]{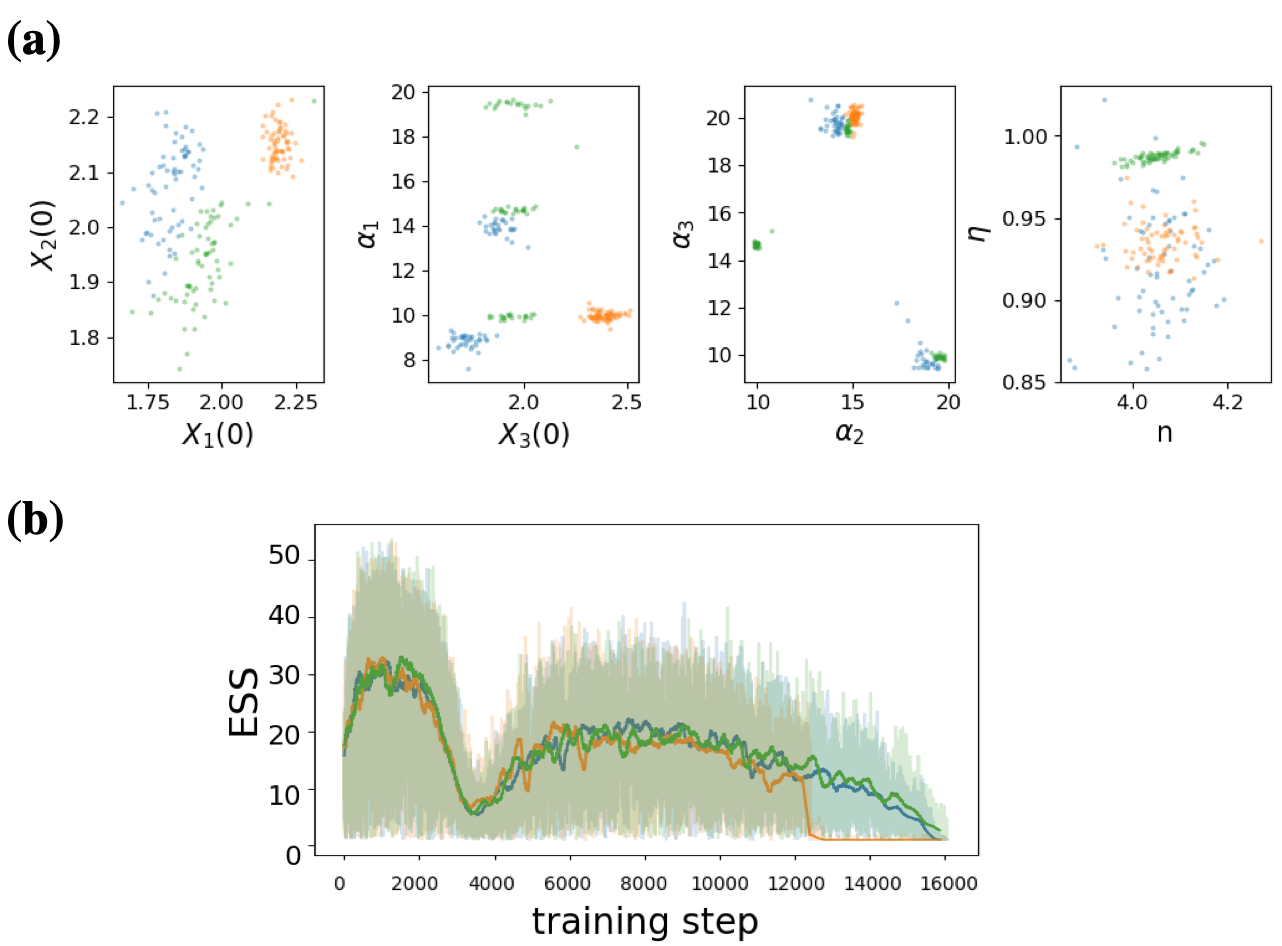}
\caption{Sampling the parameter space of the repressilator with an NF using a preset schedule for $\beta_s$: $\beta_s =( s/1000 )^4$ for $s = 1, 2, \cdots, 1000$ \cite{friel2008powerPosteriors}. 
For each $\beta_s$, the NF is trained for $800$ steps with samples updated every $50$ steps ($k=50$).
(a) Samples from NF models trained with fixed $\beta=1$. Different colors distinguish samples from three models trained with distinct initial weights and random seeds.
(b) The effective sample size (ESS). Translucent lines show the instantaneous ESS values ($n_\mathrm{eff}$) and opaque lines show their exponential moving average ($\bar{n}_\mathrm{eff}$) with $\lambda=0.01$.  Colors correspond to the runs in (a).
The drop in ESS is consistent with the NFs failing to converge.  In particular, the orange run exhibits both a sudden drop and mode collapse. 
} 
\label{fig:preset_schedule}
\end{figure}

\begin{figure}[h]
\centering
\includegraphics[width=0.8\linewidth]{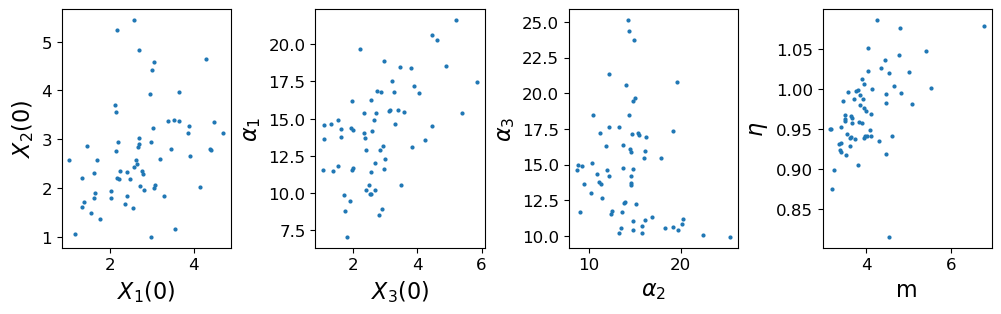} %
\caption{ Data sampled from an NF trained with ESS threshold $n^* = 0.2$. }
\label{fig:samples_ESS-threshold-0.2}
\end{figure}

\begin{figure}[h]
\centering
\includegraphics[width=0.6\linewidth]{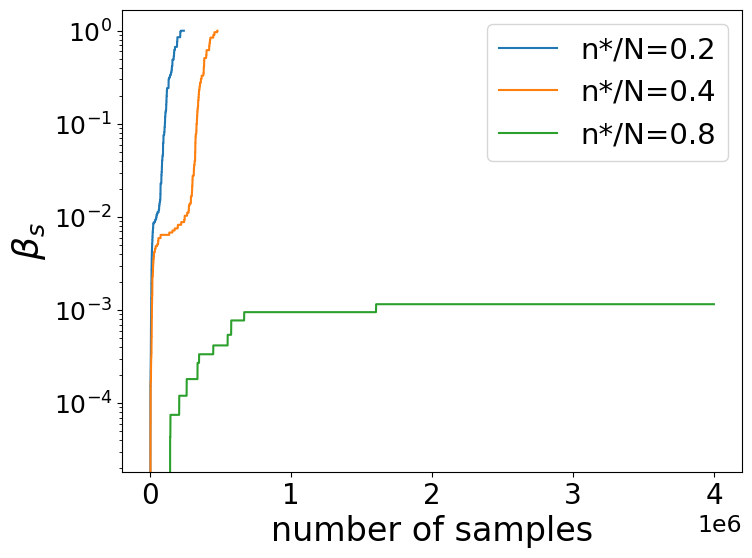} %
\caption{ The change of $\beta_s$ in training with different ESS thresholds, $n^*$. Other hyperparameters are  $L=8$ and $k=50$. }
\label{fig:compare_ESS_threshold}
\end{figure}

\begin{figure}[h]
\centering
\includegraphics[width=0.8\linewidth]{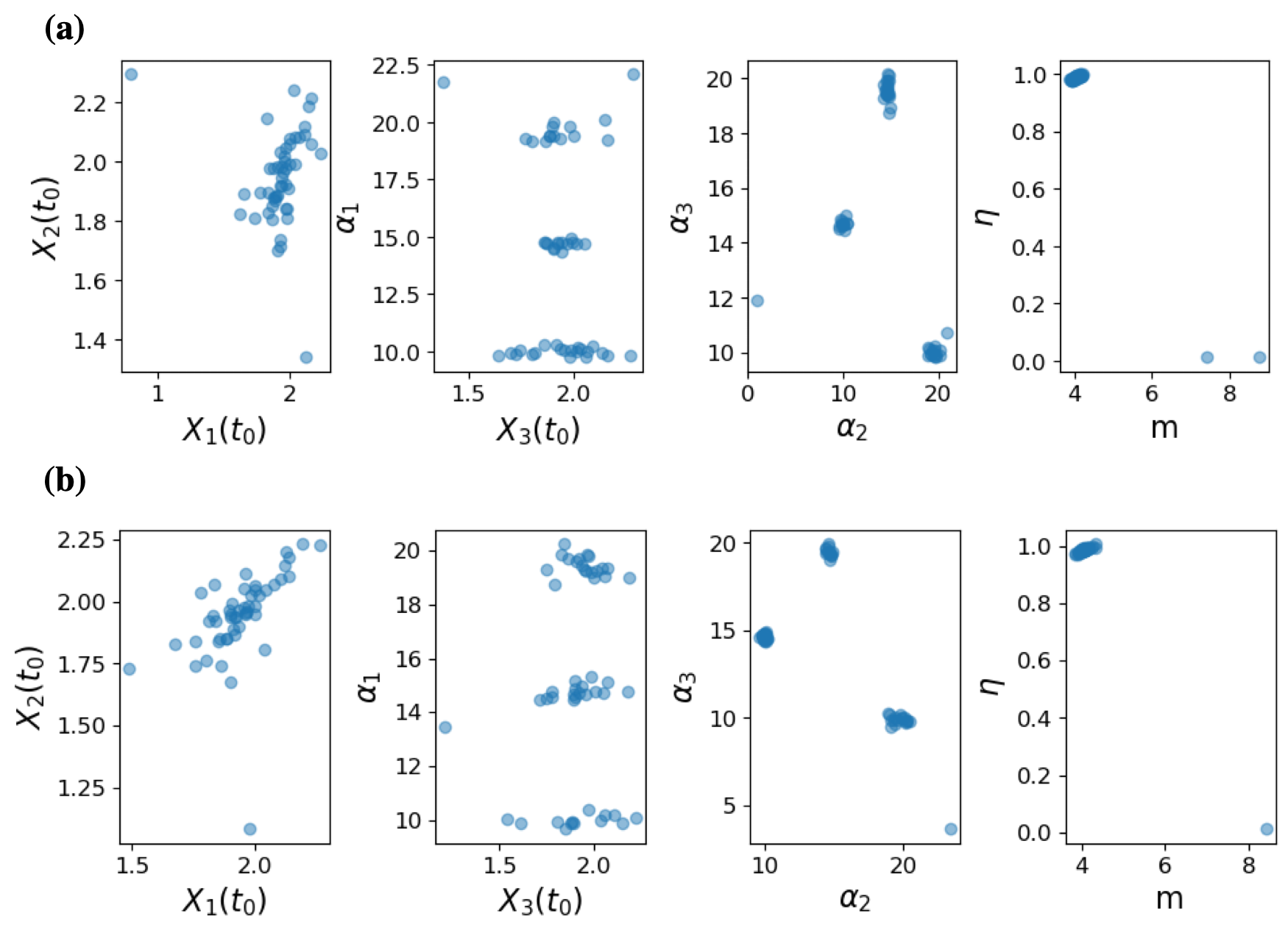} %
\caption{Results for ensemble MCMC with different annealing schedules: (a) 1000 MCMC steps for each $\beta$ and (b) 1500 MCMC steps for each $\beta$. }
\label{fig:MCMC_outliers}
\end{figure}

\begin{figure}[h]
\centering
\includegraphics[width=0.6\linewidth]{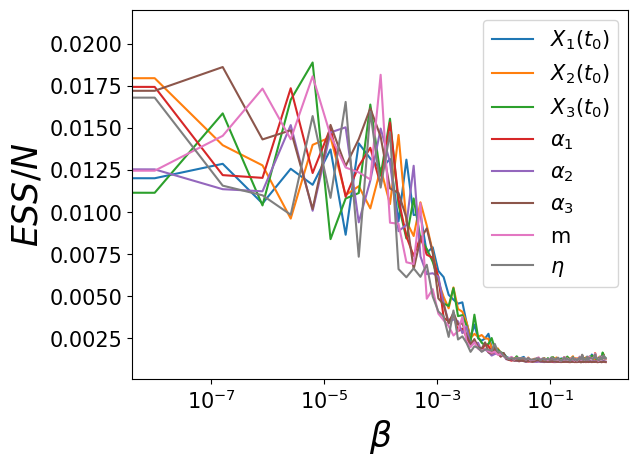} %
\caption{Univariate effective sample size (ESS) for each parameter obtained from MCMC.}
\label{fig:MCMC_ESS}
\end{figure}

\end{document}